\documentclass[a4paper,conference]{IEEEtran}
\usepackage{times}
\usepackage{epsfig}
\usepackage{graphicx}
\usepackage[outdir=./figures/]{epstopdf}
\usepackage{amsmath}
\usepackage{amssymb}
\usepackage{siunitx}
\usepackage{booktabs}

\ifCLASSINFOpdf
\else
\fi
\begin{document}

%
\title{Handwritten Character Recognition from Wearable Passive RFID}

\author{\IEEEauthorblockN{Leevi Raivio \hspace{2cm} Han He \hspace{2cm} Johanna Virkki \hspace{2cm} Heikki Huttunen} \\
\IEEEauthorblockA{Faculty of Information Technology and Communication Sciences \\
Tampere University \\
Tampere, Finland\\
Email: leevi.raivio@tuni.fi}}


%


\maketitle

\begin{abstract}
In this paper we study the recognition of handwritten characters from data captured by a novel wearable electro-textile sensor panel. The data is collected sequentially, such that we record both the stroke order and the resulting bitmap. We propose a preprocessing pipeline that fuses the sequence and bitmap representations together. The data is collected from ten subjects containing altogether 7500 characters. We also propose a convolutional neural network architecture, whose novel upsampling structure enables successful use of conventional ImageNet pretrained networks, despite the small input size of only 10x10 pixels. The proposed model reaches 72\% accuracy in experimental tests, which can be considered good accuracy for this challenging dataset.
Both the data and the model are released to the public.
\end{abstract}

%
\IEEEpeerreviewmaketitle

\section{Introduction}

Wearable technology is a growing trend both in consumer electronics and healthcare \cite{wearables_and_future, wrbls_medicide}, and new types of devices are introduced frequently. Smart clothing is among the most recent technologies in wearable devices. Progress in nanotechnology, electro-textiles and wireless technology \cite{elt_ebm_struct_for_wrbl, etextiles_rfid, RIO, elt_rfid2017} have recently enabled the production of textiles that that have electrical functionalities, as well. In particular, using wearable sensors as flexible and transparent user interfaces has a number of potential uses. Our focus is on how handwriting could be used as a modality of man-machine interaction using such devices.

Online handwriting recognition has been studied extensively in the past \cite{hwr1990, hmm1996, blindtyping, fujitsuonln, likelihoodnorm}, and has also been applied to wearable sensors \cite{airwriting, wondersense}. However, past applications have seen little use, due to the need for an integrated power source and maintenance, making them difficult to apply for a real-world scenario.

Recently, a novel technique for constructing a wireless textile touch input device based on passive sensors \cite{eltpaper} was proposed. Passive ultra-high frequency (UHF) radio frequency identification (RFID)-based solution allows natural and easy-to-use user interfaces, indistinguishably integrated into our everyday clothing. Unlike other technologies, the solution is fully maintenance-free, having no on-cloth energy sources, which makes its implementation cost-effective and simple. The cost of a passive RFID integrated circuit (IC) is only a few cents. Possible users of the platform are especially special needs users, such as people with speech and language or cognitive challenges. Compared to mobile devices, which require a certain amount of cognitive skills, they can be lost, or they may be out of the battery, this platform is simple to use and, when integrated into clothing, always with the user.

However, due to the limitations posed by the materials and the environment, the quality of the data is clearly inferior to non-textile input devices such as tablets or smartphones, and even humans struggle when attempting to categorize the resulting bitmaps. However, the data consists of both the bitmap and the order in which different parts of the symbol were drawn. In this paper, we propose a method that integrates this temporal information together with the spatial information, thus reaching superhuman accuracies. Moreover, we release the first handwritten digit dataset collected with a wearable device, which we call \textit{Wearable MNIST}. 

A major challenge with the data is that the input device is not always sensitive enough to capture all individual touch events. This is due to both the mechanical designs as well as the user environment: If the device is part of clothing, parts of the sensor matrix may be out of touch due to wrinkles in the fabric. Therefore, the proposed method should be robust against missing touch events. A part of our strategy is to interpolate between recorded data points, which makes the digit shape already easier to interpret. Moreover, we present a convolutional neural network structure that learns to cope with the imperfect data.

The classifiers are tested with real-world applications in mind, with data collected from eight subjects with different ethnicities. We experiment both with tests subjects known and unknown by the network, and discover that the handwriting style of the subjects seems to differ significantly, and the resulting accuracies at test time have a large variance. Therefore, we also consider another scenario, where new users are required to calibrate the system by drawing a small number of examples before starting to use the system. This can be simulated by collecting additional data from test subjects already represented in the training set, in a separate session. 

The remainder of this paper is structured as follows. First, we describe the design of the data capture device in Section 2, and detail the collection of data in Section 3. Next, Section 4 proposes a deep neural network structure for classification of the handwritten digits, together with a preprocessing pipeline to improve the quality of the input data. In Section 5, we experiment with the proposed method on the character dataset. Finally, Section 6 discusses the results and concludes the paper.

\begin{figure*}[t]
\begin{center}
\includegraphics[width=0.8\linewidth]{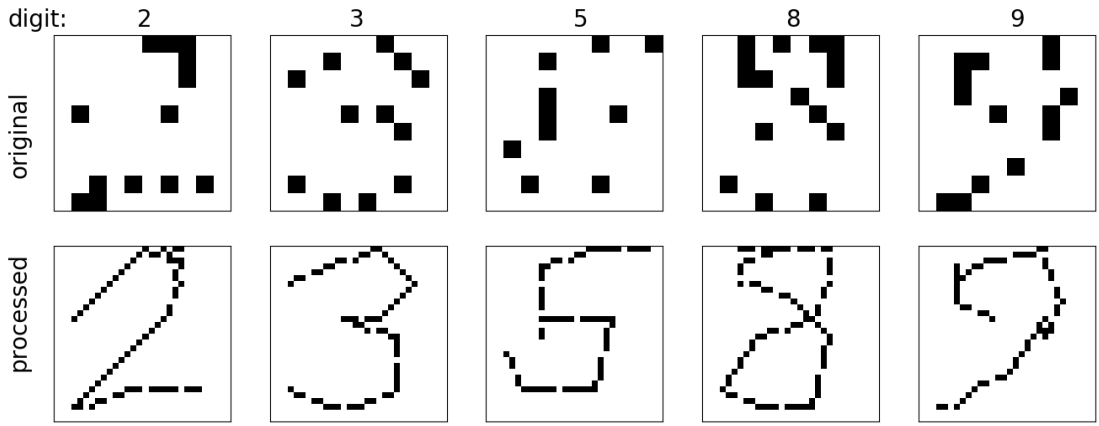}
\end{center}
   \caption{Samples from the dataset. Top row: raw data; bottom row: interpolated data.}
\label{fig:samples}
\end{figure*}

\begin{figure*}[t]
\begin{center}
\includegraphics[width=0.8\linewidth]{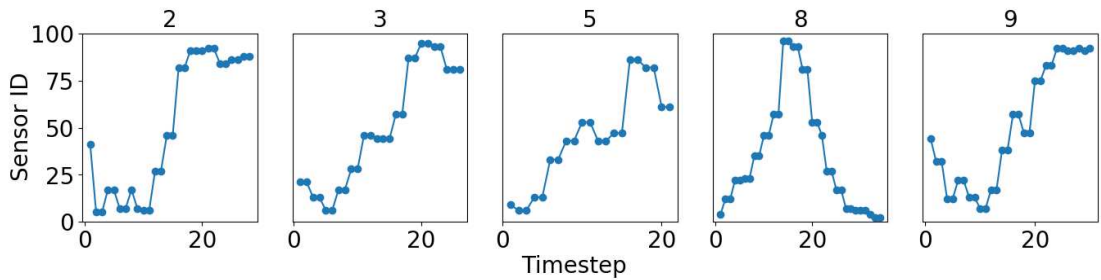}
\end{center}
   \caption{Data samples as sequences.}
\label{fig:sequences}
\end{figure*}

\section{Textile touch platform}\label{sec:platform}

\begin{figure}[t]
\centering
\includegraphics[width=\linewidth]{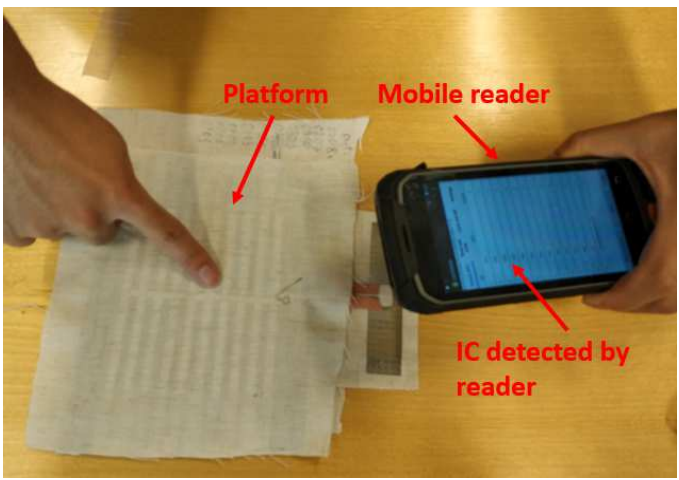}
\caption{Textile touch input device.}
\label{fig:platform}
\end{figure}

\begin{figure*}[t]
\begin{center}
\includegraphics[width=0.8\linewidth]{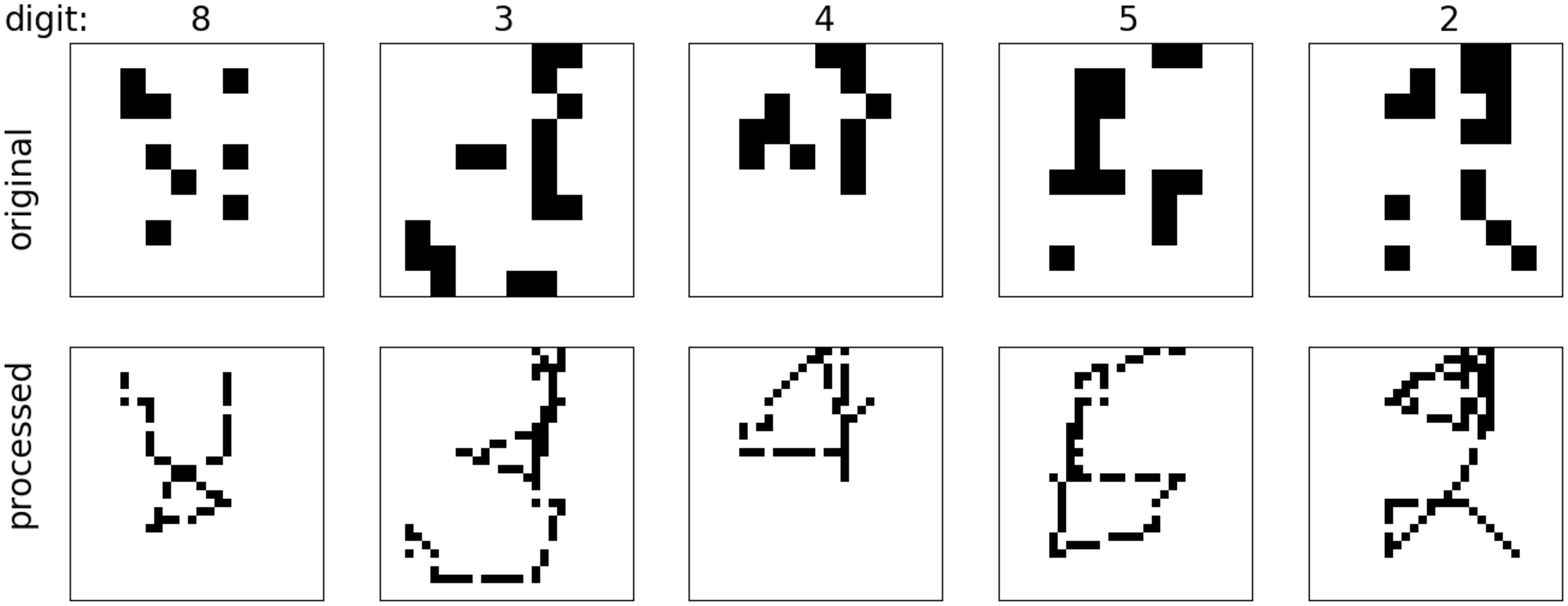}
\end{center}
   \caption{Challenging samples from the Wearable MNIST dataset.}
\label{fig:ch_samples}
\end{figure*}

The data for this paper is collected with a passive RFID (radiofrequency identification) based keyboard \cite{eltpaper}, which can be used as an input method for mobile devices. The prototype of the platform consists of two textile layers. The bottom layer includes a cotton fabric substrate and 100 attached passive RFID microchips. The used microchip, \emph{i.e.}, integrated circuit (IC) is NXP UCODE G2iL series RFID IC. On the bottom layer, the 100 ICs are arranged as a $10\times 10$ array to form a 2-dimensional matrix for drawing. Each IC is associated to its position in the array, and thus raw data will consist of a 100-dimensional time series, which can be converted to a $10\times 10$ bitmap.

The resolution in this first prototype is low, as the idea is only to test the concept with ready commercial RFID ICs, which were attached to the strap structures by the manufacturer. These ICs were selected for fast and easy fabrication of the prototype textile platform. However, we are by no means limited to the current number of ICs or the space between individual ICs.

For transmitting the measurement data in a wireless manner, the right sides of the ICs are cascaded and connected to a dipole antenna by electro-textile circuits on the bottom layer. Moreover, the left sides of the ICs are connected to the antenna by electro-textile circuits on the top layer. An elastic and flexible rubber material is utilized as a separation between the two layers to avoid them directly touching. Thus, all the ICs are initially disconnected from the antenna. When symbols are written on this platform, touch will create an electrical connection between the top and bottom layers. After that, an RFID reader can be used to detect and record the activated ICs.

The platform allows one to send real-time touch-input data wirelessly to, \textit{e.g.}, a smartphone. Since it's based on passive RFID technology, where the communication between the reader and the ICs is based on signal backscattering, an internal power source isn't needed. This makes the device better suited to be integrated into clothing, such as into a shirtsleeve. The prototype used is robust enough to endure real use. It had a stable performance during the collection of the dataset presented in Section \ref{sec:dataset}.

\section{Dataset}\label{sec:dataset}

In this chapter we describe the dataset collected with the textile touch-input device of intoduced in the previous chapter. The data consists of 7500 samples of manually labeled digits from 0 to 9, collected by eight test subjects of five different nationalities. The subjects were selected to represent different nationalities to cover different scripting styles: Finnish, Italian, Chinese, Persian and Nepalese. 

We also define a data split designed to assess the generalization over subjects in two scenarios: (1) the user starts to use the pad from scratch with no calibration of personalization, and (2) the user calibrates the system by showing examples of personal handwriting style. This way the results will indicate the significance of personalization, and whether the performance will be acceptable without it.

With this intention, the training set consists of six individual subjects, while the validation set is split in two parts consisting of (1) two new subjects unseen by the network before validation, and (2) two test subjects that have already participated in the training set. The validation and training data are collected in separate sessions on different days. Dataset details are collected to Table \ref{tab:dataset}, and examples of raw data from the captured digits can be seen the top row of Figure~\ref{fig:samples}. 

\begin{table}[!t]
\renewcommand{\arraystretch}{1.3}
\caption{Dataset statistics}
\label{tab:dataset}
\centering
\begin{tabular}{r|c|c|c}
Split                 & Training   & Known val.  & Unknown val. \\
\hline\hline
Subjects participated & 6 (\# 1-6) & 2 (\# 1, 3) & 2 (\# 7, 8)  \\
Number of samples     & 4500       & 1500        & 1500         \\
Samples per subject   & 750        & 750         & 750          \\
Samples per digit     & 75         & 75          & 75           \\
\end{tabular}
\end{table}

The examples of Figure \ref{fig:samples} illustrate the challenges of data capture. The dots are isolated due to sensitivity issues of the panel and the data collection setup. However, as we collect also the sequential order of contacts, we can connect the dots together based on their temporal order. Interpolated versions of the five digits are shown on the bottom row of Figure \ref{fig:samples}. The exact details of interpolation procedure are presented in Section \ref{subsec:preprocessing}.

Each test subject was requested to draw 75 samples of each digit with the device, totalling 750 digits per person. No more than ten samples of the same digit were written successively, to avoid users to become accustomed and develop a routine of writing the successive digits. This way, the data samples collected are more natural, and represent common use cases where users are not writing just one character multiple times in a row. The subjects weren't given any instructions on number shapes, to emphasise different natural scripting styles.
The identity of each subject was also recorded for later cross-validation use.

In addition to the $10\times 10$ binary images, the sequential raw data representation of each digit was stored. In this representation, the data consists of individual sensor activations sorted by the timestamp of the activation. Sensor id's are one-hot encoded as vectors of hundred timesteps, stacked and padded with zeros to form $100 \times 100$ two dimensional arrays. With images and sequences, both spatial (the shape of the digit) and temporal (path of the stroke) information can be interpreted. The sequential representations the same samples as in Figure \ref{fig:samples} are gathered in Figure \ref{fig:sequences}.

Because of the limitations posed by the materials and the environment, the data received from it has some flaws. Some sensors that are touched might not activate, which can make the digit hard to interpret. Some difficult examples are gathered in Figure \ref{fig:ch_samples}; with top row representing the rendered raw data and bottom row its interpolated version. Some samples are hard to read even by humans and make the classification task harder than usual handwritten character classification problems. For example, a simple convolutional neural network \cite{cnn_best_practices} that works almost perfectly for handwritten digits, \textit{e.g.} the MNIST dataset \cite{mnist}, does not work that well for this case. Also, like in other similar problems, different styles of handwriting make the digits harder to read as well, which is further emphasised by the different nationalities; both in terms of bitmap representation (bitmaps look different), and in terms of the sequential representation (subjects draw the symbols in different stroke order; for example a '0' is commonly drawn in two different rotation directions). Therefore, the classifier has to be robust for inter-class variance.

\section{Proposed method}\label{sec:solution}

\begin{figure}[t]
\begin{center}
\includegraphics[width=\linewidth]{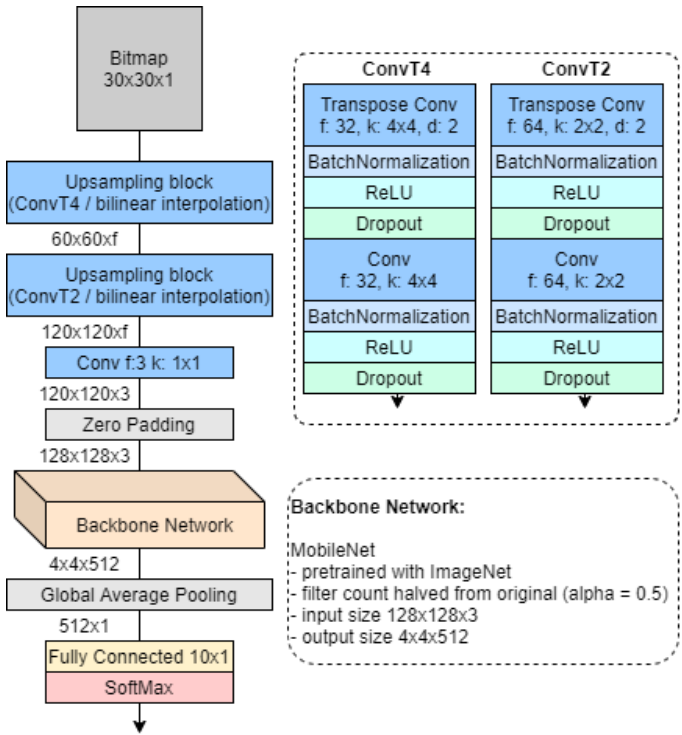}
\end{center}
   \caption{Network topography. Notation in convolutional layers is as follows: f: number of filters, k: kernel size, d: dilation. }
\label{fig:network}
\end{figure}

As stated in the previous section, the task of classifying the handwritten digit dataset collected with the textile touch platform is not trivial. In this section we propose a neural network approach for the problem. Since the data is imperfect, some samples can be hard to classify even for a human. Our solution fuses the sequential and spatial representations to reach a better accuracy than using either individually. Even though a digit might be hard to classify based on its shape alone, we can also use the trajectory of the stroke to compensate imperfections in the spatial representation. For example, digit '5' in Figure \ref{fig:ch_samples} could easily be interpreted as a '6', but their difference is in the order in which the strokes appear.

\subsection{Data preprocessing}\label{subsec:preprocessing}
As discussed earlier, some sensors in the grid might not always get triggered, resulting in imperfect bitmaps. However, the missing sensor readings can be interpolated, since the order of the sensor activations is known. Samples formatted as sequences are first interpolated, so that the empty spaces between successive activations are filled, and then images are constructed from the sequences. Examples of this are shown in the lower row of Figure \ref{fig:samples}.

We also found that by increasing the resolution of the images, classification accuracy gets better. Larger resolution means deeper convolutional neural networks can be applied, even with pooling layers included. We observed that this seems to result in the convolution layers of the classifier to distinguish the digit shapes better. In the end, the spatial resolution from the original $10\times 10$ pixels is increased to $30\times 30$ pixels.

To conclude, the preprocessing pipeline is as follows: first, raw data from the sensor is trasformed to one-hot encoded sequences. Second, the resolution of the sequences are increased three-fold, and missing samples are interpolated bilinearly. Last, $30 \times 30$ bitmaps are constructed from the interpolated sequences.

\subsection{Network Structure}\label{subsec:classifier}
We compare three different architectures for the classifier: (1) A small convolutional neural network (CNN) that receives preprocessed input data in bitmap format, (2) a deeper pretrained CNN in which the bitmaps are first upsampled with bilinear interpolation and then passed through a backbone CNN and (3) a similar network to the second approach, where the bitmaps are upsampled with transpose convolution layers instead. Model topography of the two latter approaches is presented in Figure \ref{fig:network}.

The smallest input resolution for Mobilenet with pretrained weights available is $128\times 128$. For the purposes of this paper, the full pretraining at smaller resolution would be time-consuming; and probably harmful to the network topology that are designed for higher input sizes (in terms of downsampling stages and respective receptive field sizes). Backbones pretrained with MNIST \cite{mnist} were also tried, but provided no significant improvement in accuracy compared to existing ones pretrained with ImageNet \cite{imagenet}. Therefore, we decide to adjust our input to existing pretrained networks instead of pretraining a network to match our input. In addition to the resolution, the input for the pretrained network should have three color channels, while our input is binary. Therefore, we insert upsampling layers to increase the resolution and simultaneously generate synthetic RGB channels. 

The detailed structure of the three DNN architectures is the following.  
The first, simplest architecture ("Simple CNN" in the sequel) consists of two regular CNN blocks similar to the ConvT blocks of Figure \ref{fig:network}, followed by a max pooling layer. This block is repeated twice, followed by one more convolutional layer, average pooling, fully connected layer and a softmax activation layer. 

The Mobilenet based pretrained architectures start by two upsampling layers (\textit{ConvT} blocks in Figure \ref{fig:network}---inspired by a decoder network of autoencoders and Generative Adversarial Networks \cite{GAN}). Two different approaches are tested: In the first one, the input bitmaps are simply upsampled with bilinear interpolation, and then passed through a shallow convolutional layer and padded to compose an input to the backbone network. In the second approach, transpose convolution layers \cite{convarithmetic} are utilized instead. 

While in the first approach, the features are only passed through convolution layers and downsampled, in the two latter approaches the features are first upsampled before passing them through a similar, although much deeper, structure. Upsampling is assumed to raise accuracy, since it allows one to better analyze the features in multiple scales, and enables the use of a much deeper backbone. Transpose convolution layers might also find some additional features that have a positive effect on accuracy, and further clean up the imperfections in the data.

\section{Experimental results}\label{sec:tests}

The three proposed architectures are experimented with two different end-user scenarios. First, we assume that the device uses generic training data and the system has not been calibrated with user-specific data, and thus the validation set with two test subjects previously unknown to the network is used. In the second scenario, we assume that the end-user first calibrates the system with a small amount of their own handwriting samples. We simulate this scenario by using the validation set collected by two test subjects already appearing in the training set. 

Training setup is identical for all tests.
To make the results reproducible, and to ensure fair evaluation, the random seed is set at a certain fixed value. The networks are trained with stochastic gradient descent. Learning rate is initially set to 0.01, and reduced when validation accuracy reaches a plateau. The networks are trained until validation accuracy no longer increases. Minibatch size is 32 samples.

\subsection{Effect of different preprocessing strategies}

We study the different preprocessing options using the simple CNN classifier due to its fast training speed. More specifically, we are interested in the effect of the temporal interpolation, input resolution and data augmentation. To this aim, the simple CNN classifier (see Section \ref{subsec:classifier}) is trained using different preprocessing configurations while retaining all other network hyperparameters the same (due to global average pooling, the identical convolutional pipeline can be used for all input resolutions). The accuracy is  in this case evaluated by 6-fold leave-one-subject-out crossvalidation leaving one of the six training subjects out at a time. Results are gathered in Table \ref{tab:preprocessing}.

\begin{table}[tbh]
\caption{The effect of preprocessing on CNN accuracy.}
\begin{center}
\begin{tabular}{r||cc}
Input resolution           & $10\times 10$  & $30\times 30$ \\
\hline\hline
Raw pixel data & 0.53 & 0.62  \\
Temporal bilinear interpolation & 0.56 & 0.69 \\
Interpolation \& augmentation & 0.59 & \textbf{0.73} \\
\end{tabular}
\end{center}
\label{tab:preprocessing}
\end{table}

\begin{table*}[t]
\caption{Classification accuracies for four test subjects. Subjects 1 and 2 ("known") have also drawn additional 750-example sets of samples to the training set, while Subjects 3 and 4 ("unknown") have not.}
\label{tab:networkacc}
\begin{center}
\begin{tabular}{l||c|c|c||c|c|c}
 & \multicolumn{3}{c||}{Known} & \multicolumn{3}{c}{Unknown}              \\ 
 Classification model \textbackslash Test set partition                      & Subject 1    & Subject 2     & Mean        & Subject 3    & Subject 4    & Mean          \\
\hline\hline
Simple CNN (51k params)                    & 0.533 & 0.395 & 0.464       & 0.704 & 0.675 & 0.690               \\
MobileNet \& bilinear upsampling (814k params) & 0.748 & 0.610 & 0.679       & 0.763 & 0.679 & 0.721                 \\
MobileNet \& ConvT upsampling (866k params) & \bf{0.772} & \bf{0.617} & \bf{0.695}  &\bf{0.785} & \bf{0.696} & \bf{0.741}\\

\end{tabular}
\end{center}
\end{table*}

From the results, one can see that the increase in resolution has the largest effect in increasing the accuracy. Larger image resolutions lead to more features to process, since image dimensions are reduced in the pooling layers of the network. Even with raw data, the accuracy increases by almost 10 \%-points, and even more with the interpolation and augmentation added.
Moreover, the interpolation between temporally successive measurements improves the classifier accuracy even in the original $10\times 10$ resolution but even more clearly with the larger $30\times 30$ input size. Despite the information content remains the same, the temporal interpolation transforms the representation into a more suitable form for the classifier to solve, although it would technically be possible for it to learn a similar way of transforming the data. 

Data augmentation added after the temporal interpolation increases the accuracy even further. Our augmentation steps consist of a random rotation (max ten degrees) and random translation (max two pixels). Since the best accuracies were reached with interpolation, resolution increase and augmentation, these are applied in the following tests, as well.

\subsection{Network architectures and inter-subject generalization}\label{subsec:test_clf}

The results for both test scenarios for each network architecture are shown in Table \ref{tab:networkacc}, where we compare the simple CNN  architecture trained from scratch with Mobilenet \cite{mobilenet} network pretrained with Imagenet dataset \cite{deng2009imagenet}. However, as described in Section \ref{subsec:classifier}, the smallest input resolution for Mobilenet with pretrained weights available is $128\times 128$. In addition, the input for the pretrained network should have three color channels, while our input is binary at resolution $30\times 30$. Therefore, we insert upsampling layers to increase the resolution and simultaneously generate synthetic RGB channels. 

On the other hand, Table  \ref{tab:networkacc} studies the generalization of the models across subjects. More specifically, the two types of test subjects---two subjects with another set of samples in the training set and two "fresh" subjects without training examples---should reveal whether calibration with user-specific data is beneficial.

Regarding the accuracy of different models, it can be seen from Table \ref{tab:networkacc} that the simple CNN architecture is clearly inferior to the deeper architectures, with up to one third more errors than the deeper pretrained models.
Moreover the transpose convolution layers in the upsampling pyramid increases accuracy by a few per cent. Although the improvement brought by  the upsampling pyramid over a bilinear upsampling operation is limited, it is consistent across all test subjects.

\begin{figure*}[t]
\begin{center}
\includegraphics[width=0.8\linewidth]{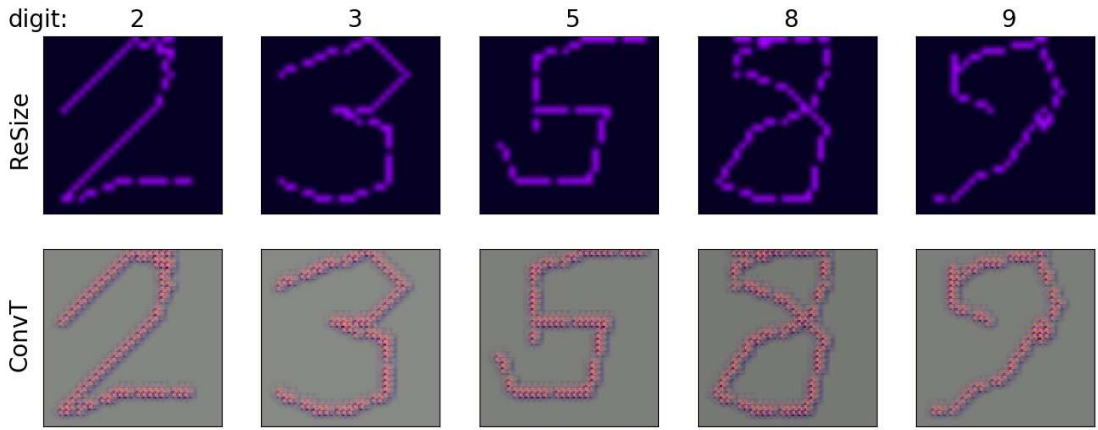}
\end{center}
\caption{Results of the two alternative upsampling operations for all samples of Figure \ref{fig:samples} as RGB images.}
\label{fig:sample_activations}
\end{figure*}

In order to further study the effect of the upsampling pyramid, we illustrate it's outputs,  \textit{i.e.} the backbones inputs, in Figure \ref{fig:sample_activations} for the five digits of Figure \ref{fig:samples}. Here, the inputs of the upsampling pyramid are the $30\times 30$ binary images and the outputs are $128\times 128$ RGB images further fed to the Mobilenet for classification. Moreover, the figure shows also the result of conventional bilinear upsampling operation (top). The activations from the transpose convolution pyramid are a bit smoother and are spread more evenly across all three colour channels.
We believe that the increased accuracy due to the transpose convolution layers is because they learn to compensate for the imperfections in the data better.

\begin{table}[h]
\caption{Accuracies on subjects S1 and S2 with original data and a variant where S1/S2 are replaced by another subject. Classification model was MobileNet with ConvT upsampling.}
\label{tab:uncalibrated_known_set}
\begin{center}
\begin{tabular}{r|c|c}
Data                          & S1 replaced & S2 replaced \\
\hline\hline
Original          & 0.772      & 0.617       \\
Test subject removed from calibration  & 0.678      & 0.532       \\
\end{tabular}
\end{center}
\end{table}

There is significant variation in individual subject's accuracies. This verifies the assumption that the classification is sensitive to different scripting styles. Due to this fact it is also impossible to decide whether the assumption that the network performs better with test subjects it has seen before holds. The mean accuracies of known subjects are smaller than the ones with unknown subjects, but more test subjects would be needed to draw any conclusions. 

However, in order to assess the significance of calibration, we can remove the effect of calibration by replacing Subject 1 or Subject 2 in the training set by new data. For the new data we decide to use that of test Subject 3, which is not contained in training set. The results of this experiment are described in Table \ref{tab:uncalibrated_known_set}. It can be seen that the accuracy drops by up to 10\% points. This indicates that calibration indeed does increase classification accuracy.

Possible ways of coping with errors related to different styles of handwriting and thus reducing variance between test subjects could be to design a network that is less sensitive to these errors, or by specifying a certain handwriting style to use with the device. Even though the problem of different handwriting styles would be eliminated with a fixed style, the uncertainties related to different ways of using the device would still remain. However, as we discover in the following section, users tend to automatically adapt to the device as well, reducing uncertainties significantly.

\subsection{Live tests}\label{subsec:test_live}

\begin{figure}[t]
\begin{center}
\includegraphics[width=\linewidth]{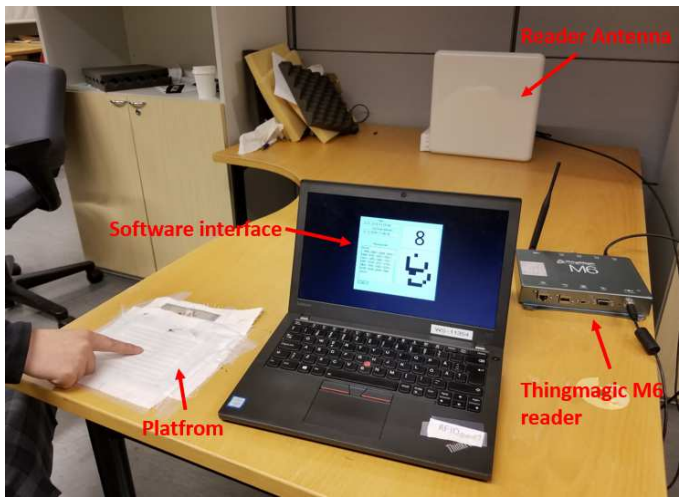}
\end{center}
   \caption{Test setup for live character recognition in \cite{eltpaper2}.}
\label{fig:live_setup}
\end{figure}

In \cite{eltpaper2}, the accuracy of the simple CNN classification model was tested in practice. In addition to the platform and an RFID reader, the setup also consists of a laptop with a graphical user interface, allowing the users to draw numbers and see their recognition results, as well as the bitmaps they drew. The test setup is presented in Figure \ref{fig:live_setup}.

Six test subjects participated in the test session, four of whom had never used the device before. Two of the test subjects had experience with the device through collecting dataset presented earlier. The test were conducted in an office, with normal distractions and noise sources, \textit{e.g.} people talking, WiFi, and phone signals. The users were allowed to practice before the actual tests, so that their performance  wouldn't significantly increase during the test.

\begin{figure}[t]
\begin{center}
\includegraphics[width=\linewidth]{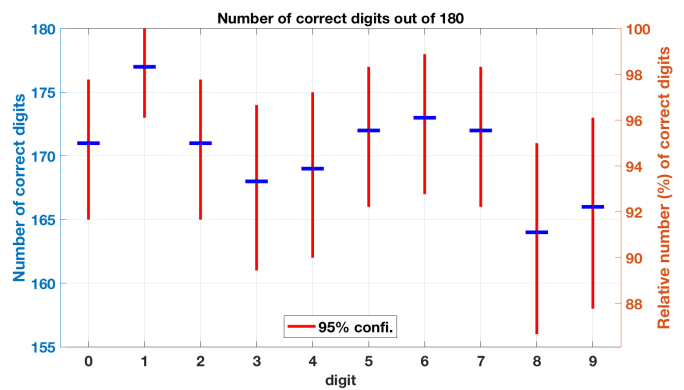}
\end{center}
   \caption{Results of live character recognition for each digit in \cite{eltpaper2}.}
\label{fig:live_results}
\end{figure}

The live test results proved to be much better than in the offline tests: mean accuracy across all test subjects was $94.6 \%$---far beyond any accuracies reached in the experimental results with static Wearable MNIST data. Accuracies and confidence intervals for individual digits are shown in Figure \ref{fig:live_results}. We believe this increase in accuracy is because of the immediate feedback the user gets from the screen: the user interface gives the user feedback on how well their handwriting is classified, which results in the user learning to write in a way that gets better results. This is in line with many other modes of human-computer interaction: for example, people automatically speak more clearly even to the most up to date speech recognition systems. Moreover, as these results were achieved with the simplest CNN presented, one could expect a few percent improvement if one of the deeper architectures were used.

Apart from the high average, Figure \ref{fig:live_results} shows also a large variation in accuracy across digits; for example, digit '8' is the most challenging to categorize while 2 is the easiest. Also, the within-class variance is significantly larger for some classes. We believe this to be the cause of different scripting styles and handwritings: some digits are written in much larger range of styles than others. Also, users might have different styles of using the device itself. Therefore, accuracy could be greatly increased by designing an approach that would learn to interpret different styles of writing better, for example by using the sequential information as an input to the network as well, thus fusing the spatial and temporal data together in a data-driven manner.

\section{Conclusion}

The objective of this work was to present a solution to handwritten digit recognition with a wearable touchpad. While the task might seem easy on the surface, imperfections in the data due to mechanical designs and user environment make it a harder problem than character recognition with non-textile devices. As a solution, we proposed two ways of combining spatial and sequential information to increase classification accuracy on this device. Interpolating between data points, increasing bitmap resolution and data augmentation proved to increase performance greatly. A deep neural network structure with an upsampling pipeline and a backbone network was also proposed.

Tests revealed that upsampling the input features before passing them through a backbone CNN seem to increase classification accuracy, as well as using transpose convolution layers instead of resizing with bilinear interpolation. The solution seems quite sensitive to uncertainties related to an individual's handwriting style, as well as their style of using the device. Thus, future research should be aimed towards both reducing these uncertainties as well as the sensitivity related to those. With enough data representing different ethnicities and scripting styles more thoroughly, these uncertainties might decrease enough to not matter as much. In production environment, one would gather a much larger dataset. A fixed type of scripting style would also have made the problem easier. However, it would restrict the user more. Therefore, intentionally, to represent a more natural user-base, the test subjects weren't given instructions on digit shapes. It was discovered in live tests, however, that by giving the user graphical feedback on their handwritigs classification accuracy, they learn to write in a way much easier for the classifier to interpret.

In our future work we plan to collect more data and extend the dataset to other characters. Different designs for integrating the temporal and spatial data representations together shall also be tested. Integration of LSTM and GRU networks with the CNN have been tested to combine spatial and temporal representations, with no significant improvement so far. For simplicity, we have thus omitted these structures from this work but will definitely be studying this aspect more in the future. Temporal and spatial information can be fused in different phases of the classification pipeline, and they might have the additional benefit of also learning functions similar to current preprocessing steps as well. Additionally, as textile touch-interfaces develop, more accurate and consistent data could be acquired. 

Even this early prototype of the wearable device provides robust and precise enough data to enable new modality for man-machine interaction. It seems that wearable devices---including smart clothing---continue to gain popularity. Therefore, integrating user-interfaces to ones clothing could be the next step in the development of consumer electronics. 
However, a crucial bottleneck in their adoption is how the data can be interpreted in a reliable and robust manner. If this problem were solved, the development in both textile electronics and data processing related to them will provide increasingly impressive wearable devices.

\bibliographystyle{IEEEtran}
\bibliography{refs}

\begin{thebibliography}{10}
\providecommand{\url}[1]{#1}
\csname url@samestyle\endcsname
\providecommand{\newblock}{\relax}
\providecommand{\bibinfo}[2]{#2}
\providecommand{\BIBentrySTDinterwordspacing}{\spaceskip=0pt\relax}
\providecommand{\BIBentryALTinterwordstretchfactor}{4}
\providecommand{\BIBentryALTinterwordspacing}{\spaceskip=\fontdimen2\font plus
\BIBentryALTinterwordstretchfactor\fontdimen3\font minus
  \fontdimen4\font\relax}
\providecommand{\BIBforeignlanguage}[2]{{%
\expandafter\ifx\csname l@#1\endcsname\relax
\typeout{** WARNING: IEEEtran.bst: No hyphenation pattern has been}%
\typeout{** loaded for the language `#1'. Using the pattern for}%
\typeout{** the default language instead.}%
\else
\language=\csname l@#1\endcsname
\fi
#2}}
\providecommand{\BIBdecl}{\relax}
\BIBdecl

\bibitem{wearables_and_future}
M.~{\c C}i{\c c}ek, ``Wearable technologies and its future applications,''
  \emph{International Journal of Electrical, Electronics and Data
  Communication}, vol.~3, pp. 2320--2084, 05 2015.

\bibitem{wrbls_medicide}
A.~K. Yetisen, J.~L. Martinez-Hurtado, B.~Ünal, A.~Khademhosseini, and
  H.~Butt, ``Wearables in medicine,'' \emph{Advanced Materials}, vol.~30,
  no.~33, p. 1706910, 2018.

\bibitem{elt_ebm_struct_for_wrbl}
H.~He, X.~Chen, L.~Ukkonen, and J.~Virkki,
  ``\BIBforeignlanguage{English}{Textile-integrated three-dimensional printed
  and embroidered structures for wearable wireless platforms},''
  \emph{\BIBforeignlanguage{English}{Textile Research Journal}}, vol.~89,
  no.~4, 2019.

\bibitem{etextiles_rfid}
K.~{Koski}, L.~{Syd{\"a}nheimo}, Y.~{Rahmat-Samii}, and L.~{Ukkonen},
  ``Fundamental characteristics of electro-textiles in wearable uhf rfid patch
  antennas for body-centric sensing systems,'' \emph{IEEE Transactions on
  Antennas and Propagation}, vol.~62, no.~12, pp. 6454--6462, Dec 2014.

\bibitem{RIO}
S.~Pradhan, E.~Chai, K.~Sundaresan, L.~Qiu, M.~A. Khojastepour, and
  S.~Rangarajan, ``Rio: A pervasive rfid-based touch gesture interface,'' in
  \emph{Proceedings of the 23rd Annual International Conference on Mobile
  Computing and Networking}, ser. MobiCom '17.\hskip 1em plus 0.5em minus
  0.4em\relax New York, NY, USA: ACM, 2017, pp. 261--274.

\bibitem{elt_rfid2017}
J.~Virkki, Z.~Wei, A.~Liu, L.~Ukkonen, and T.~Bj{\"o}rninen,
  ``\BIBforeignlanguage{English}{Wearable passive e-textile uhf rfid tag based
  on a slotted patch antenna with sewn ground and microchip
  interconnections},'' \emph{\BIBforeignlanguage{English}{International Journal
  of Antennas and Propagation}}, vol. 2017, 2017.

\bibitem{hwr1990}
C.~C. {Tappert}, C.~Y. {Suen}, and T.~{Wakahara}, ``The state of the art in
  online handwriting recognition,'' \emph{IEEE Transactions on Pattern Analysis
  and Machine Intelligence}, vol.~12, no.~8, pp. 787--808, 1990.

\bibitem{hmm1996}
{Jianying Hu}, M.~K. {Brown}, and W.~{Turin}, ``Hmm based online handwriting
  recognition,'' \emph{IEEE Transactions on Pattern Analysis and Machine
  Intelligence}, vol.~18, no.~10, pp. 1039--1045, 1996.

\bibitem{blindtyping}
J.~Tokuno, N.~Akira, M.~Nakai, H.~Shimodaira, and S.~Sagayama,
  ``Blind-handwriting interface for wearable computing,'' 01 2003.

\bibitem{fujitsuonln}
H.~Tanaka, N.~Iwayama, and K.~Akiyama, ``Online handwriting recognition
  technology and its applications,'' vol.~40, pp. 170--178, 07 2004.

\bibitem{likelihoodnorm}
O.~{Velek}, S.~{Jaeger}, and M.~{Nakagawa}, ``A new warping technique for
  normalizing likelihood of multiple classifiers and its effectiveness in
  combined on-line/off-line japanese character recognition,'' in
  \emph{Proceedings Eighth International Workshop on Frontiers in Handwriting
  Recognition}, 2002, pp. 177--182.

\bibitem{airwriting}
C.~Amma, M.~Georgi, and T.~Schultz, ``Airwriting: A wearable handwriting
  recognition system,'' \emph{Personal and Ubiquitous Computing}, vol.~18, 01
  2014.

\bibitem{wondersense}
L.~{Jing}, Z.~{Dai}, and Y.~{Zhou}, ``Wearable handwriting recognition with an
  inertial sensor on a finger nail,'' in \emph{2017 14th IAPR International
  Conference on Document Analysis and Recognition (ICDAR)}, vol.~01, 2017, pp.
  1330--1337.

\bibitem{eltpaper}
H.~He, X.~Chen, L.~Raivio, H.~Huttunen, and J.~Virkki, ``Passive rfid-based
  textile touchpad,'' 2019, accepted in EuCAP, Copenhagen, Denmark, 2020.

\bibitem{cnn_best_practices}
\BIBentryALTinterwordspacing
P.~Y. Simard, D.~Steinkraus, and J.~Platt, ``Best practices for convolutional
  neural networks applied to visual document analysis.''\hskip 1em plus 0.5em
  minus 0.4em\relax Institute of Electrical and Electronics Engineers, Inc.,
  August 2003. [Online]. Available:
  \url{https://www.microsoft.com/en-us/research/publication/best-practices-for-convolutional-neural-networks-applied-to-visual-document-analysis/}
\BIBentrySTDinterwordspacing

\bibitem{mnist}
\BIBentryALTinterwordspacing
Y.~LeCun and C.~Cortes, ``{MNIST} handwritten digit database,'' 2010. [Online].
  Available: \url{http://yann.lecun.com/exdb/mnist/}
\BIBentrySTDinterwordspacing

\bibitem{imagenet}
J.~Deng, W.~Dong, R.~Socher, L.-J. Li, K.~Li, and L.~Fei-Fei, ``{ImageNet: A
  Large-Scale Hierarchical Image Database},'' in \emph{CVPR09}, 2009.

\bibitem{GAN}
I.~J. Goodfellow, J.~Pouget-Abadie, M.~Mirza, B.~Xu, D.~Warde-Farley, S.~Ozair,
  A.~Courville, and Y.~Bengio, ``Generative adversarial networks,'' in
  \emph{Proceedings of the 27th International Conference on Neural Information
  Processing Systems - Volume 2}, ser. NIPS’14.\hskip 1em plus 0.5em minus
  0.4em\relax Cambridge, MA, USA: MIT Press, 2014, p. 2672–2680.

\bibitem{convarithmetic}
V.~Dumoulin and F.~Visin, ``A guide to convolution arithmetic for deep
  learning,'' 2016.

\bibitem{mobilenet}
A.~G. Howard, M.~Zhu, B.~Chen, D.~Kalenichenko, W.~Wang, T.~Weyand,
  M.~Andreetto, and H.~Adam, ``Mobilenets: Efficient convolutional neural
  networks for mobile vision applications,'' 2017.

\bibitem{deng2009imagenet}
J.~Deng, W.~Dong, R.~Socher, L.-J. Li, K.~Li, and L.~Fei-Fei, ``Imagenet: A
  large-scale hierarchical image database,'' in \emph{2009 IEEE conference on
  computer vision and pattern recognition}.\hskip 1em plus 0.5em minus
  0.4em\relax Ieee, 2009, pp. 248--255.

\bibitem{eltpaper2}
H.~He, X.~Chen, A.~Mehmood, L.~Raivio, H.~Huttunen, P.~Raumonen, and J.~Virkki,
  ``Clothface: A batteryless rfid-based textile platform for handwriting
  recognition,'' 2020, submitted to IEEE IoT journal.

\end{thebibliography}
%

\end{document}